# Design and Embedded Validation of Compact ML Models for Affective Touch Classification in a Soft Interactive Companion

Aleksandrs Vaļiševskis[a,*], Aleksandrs Okss[a], Inese Tīģere[c], Aleksejs Kataševs[b], Dina Bethere[c], Anete Hofmane[d], Airisa Šteinberga[d], Undīne Gavriļenko[c], Santa Meļķe[c], Lucie Matoušková[e]

[a] Institute of Architecture and Design, Riga Technical University, 6 Kipsalas Street, LV-1048 Riga, Latvia
[b] Institute of Mechanical and Biomedical Engineering, Riga Technical University, 6 Kipsalas Street, LV-1048 Riga, Latvia
[c] Center for Pedagogy and Social Work, Riga Technical University Liepaja Academy, 14 Liela Street, LV-3401 Liepaja, Latvia
[d] Institute of Digital Humanities, Riga Technical University, 12/1 Azenes Street, LV-1048 Riga, Latvia
[e] Masaryk University, Zerotinovo nam. 617/9, 601 77 Brno, Czech Republic
[*] **Author to whom correspondence should be addressed:** Aleksandrs Vališevskis, Aleksandrs.Valisevskis@rtu.lv.

## Abstract

Soft plush companions provide a safe and intuitive platform for affective human–robot interaction, but their deformable structure and distributed tactile signals make reliable gesture recognition difficult. This study presents a complete workflow for developing and validating compact affective-touch classifiers for an interactive plush companion. A newly collected dataset comprised 1,326 labelled recordings before curation, including interactions from 25 children, teenagers, and adults. Each classifier received 2.5-s windows containing ten capacitive channels and one accelerometer-magnitude channel. MATLAB supported acquisition, quality control, window generation, and a 468-run exploratory study of dilated one-dimensional convolutional neural networks (1D CNNs). A closely matched Python workflow then preserved participant provenance, fitted preprocessing inside each fold, and evaluated shortlisted models by 25-fold leave-one-subject-out cross-validation (LOSO-CV). On the operational 10-class task, the compact dilated CNN achieved 81.96% mean macro-F1 and 85.42% mean accuracy. The depthwise-separable CNN achieved the strongest neural result (84.50% macro-F1), whereas a linear support-vector machine using 66 predefined time-domain features achieved the best overall result (87.98% macro-F1 and 91.05% mean accuracy). Paired fold analysis showed that the linear support-vector machine and constrained random forest outperformed the compact dilated CNN, whereas differences among the tested neural models were not statistically significant after correction. Direct measurements on a 240-MHz ESP32-S3 confirmed valid deployment of the dilated CNN, depthwise-separable CNN, temporal convolutional network, linear support-vector machine, and random forest; the linear model required a 3.2 kB serialized payload and 1.405 ms mean end-to-end classifier time. A compact Transformer remained a valid predictive comparator but failed the required host–device parity test. The results show that the original dilated CNN is a viable raw-window, feature-agnostic embedded classifier, while compact physically meaningful features provide the strongest accuracy–latency trade-off for the present dataset.

## 1. Introduction

Affective touch recognition is an important direction in socially assistive robotics (SAR), education, and rehabilitation [1]. Touch is a direct and emotionally meaningful communication channel [2,3] and may complement visual and auditory interaction for children and people with autism spectrum disorder (ASD), for whom visual attention [4,5] or auditory processing [6] can differ from neurotypical patterns. An interactive system that can interpret tactile behaviour locally may therefore support more natural and private physical interaction.

Established SAR platforms such as NAO, Pepper, KASPAR, PARO, and robotic companion pets have demonstrated value in therapy, education, and eldercare [7–15]. Nevertheless, rigid mechanisms can limit the safety and authenticity of sustained physical contact. Soft sensorized companions allow users to stroke, hold, scratch, or squeeze a compliant object, but deformation, distributed sensing, and participant-specific execution make the resulting signals difficult to interpret. This is especially relevant because automatic emotion-related recognition is already challenging [16,17], and the deployed system must generalize to people who were not represented during model development.

The present work extends previously developed sensorized plush companions for educational and therapeutic use [18]. The legacy toy used an ESP32-WROVER-E and threshold-based rules to map touch to positive or negative feedback. Those rules remained effective for abrupt, high-amplitude events, but they reacted inconsistently to subtle, sustained, or multi-sensor contacts. This limitation was important in the broader interaction context, where the design and consequences of robot feedback require particular care [19–23]. The current platform therefore moves to an ESP32-S3, whose vector instructions and memory configuration are more suitable for embedded machine learning, and introduces learned tactile-gesture recognition [24].

The aim is to design, compare, and validate compact classifiers that operate on the toy's multichannel time series and can execute autonomously on the ESP32-S3. The study contributes (i) a newly collected FAIR dataset, (ii) an open MATLAB workflow for acquisition, inspection, preprocessing, and broad 1D CNN exploration, (iii) a compatible Python workflow for participant-aware evaluation and deployment artifact generation, (iv) strict subject-independent comparison of shortlisted dilated CNNs, lightweight neural alternatives, and embedded-feasible traditional models, and (v) direct target-hardware measurements. The study evaluates tactile gesture classes with affective relevance; it does not claim to infer a participant's internal emotional state or establish clinical effectiveness.

## 2. Related Works

Research in affective computing has progressed from visual and auditory affect analysis [32–35] toward physical interaction and tactile sensing [36–39]. The following review retains the application and technical context needed to position a soft, locally intelligent companion without attempting an exhaustive survey of every recognition model.

### 2.1. Socially Assistive Robots and Human–Robot Interaction

SAR systems provide cognitive or emotional assistance through social interaction rather than physical task execution [35,40]. KASPAR, NAO, PARO, and related systems have supported engagement, imitation, and emotional expression in autism, dementia, depression, and eldercare contexts [1,13,15,41,42]. Their benefits coexist with design constraints: robots used by children must tolerate rough treatment [20–22,43,44], and rigid platforms are less suited to prolonged whole-body contact than compliant companions.

Soft-bodied devices shift the interaction emphasis from visual anthropomorphism to safe tactile engagement. They can embed distributed sensors beneath a textile surface and provide device-initiated or responsive touch [23,38,39,45]. The present plush platform follows this trajectory: tactile input is the principal interaction modality, while local classification is intended to support transparent and consistent feedback rather than simulate a complete social agent.

### 2.2. Affective Touch and Emotion Classification

Touch communicates comfort, attachment, attention, and discomfort [2,3,25–27]. Capacitive, piezoresistive, and other tactile arrays have therefore been used to recognize social-touch actions [28–31,46,47]. These studies show that model choice depends on sensor configuration, label granularity, available training subjects, and computational constraints. The labels used here denote observable interaction classes with affective relevance to the toy's response policy, not direct measurements of latent emotion.

One-dimensional CNNs are effective for biosignals and multichannel sequences because local filters can learn temporal features with bounded computation [48,49,59,60]. Dilation expands the receptive field without a proportional increase in parameters [50], and related separable or bottleneck structures reduce embedded cost [62,63]. Recurrent networks remain relevant to sequence learning [61], but temporal convolutional networks (TCNs) can provide long receptive fields with parallel computation [66]. Accordingly, the bounded comparison in this study includes a dilated 1D CNN, a depthwise-separable CNN (DS-CNN), a compact TCN, and a resource-matched compact Transformer, together with a linear support-vector machine (SVM) and constrained random forest on fixed time-domain features.

Multimodal affect systems can combine touch with facial or other signals [51], but that broader fusion problem is outside the scope of a self-contained plush sensor module. The present comparison is intentionally restricted to models that can be represented and timed on an ESP32-S3; it is not intended to rank all possible temporal architectures.

### 2.3. Therapeutic and Educational Applications

Robot-assisted activities can support social engagement and emotional regulation in autism-focused settings [1,52], while companion robots have also been associated with reduced loneliness or distress in older adults [14,15]. The authors' first-generation plush companions were previously examined as therapist-mediated materials in educational and rehabilitation sessions [18,53]. That companion work addresses methodology, usability, and interaction outcomes; the current paper is limited to sensing, classifier development, subject-independent evaluation, and embedded execution.

### 2.4. The Goal of the Study

The technical goal is to replace or complement brittle threshold logic with an evidence-based classifier that recognizes tactile gestures from previously unseen users and executes within the resource budget of the current ESP32-S3 toy. The original contribution remains the design and broad evaluation of compact dilated 1D CNNs; the revised analysis places those models in a fair participant-aware comparison and verifies the practical deployment trade-offs.

The supporting contributions are the public dataset [54], the original Processing/MATLAB software pipeline [55,58], and the transition to a Python-to-C++ workflow that preserves the MATLAB processing logic while enabling leakage-controlled experiments and direct ESP32-S3 export. The dataset and software are published in accordance with FAIR principles [57].

## 3. Materials and Methods

This section describes the sensing platform, data collection, MATLAB and Python workflows, compared classifiers, participant-aware experimental design, evaluation measures, qualitative simulation, and ESP32-S3 benchmark protocol.

### 3.1. Interactive Plush Platform and Sensor System

The experimental platform was a soft plush companion containing distributed capacitive sensors beneath the textile surface and a three-axis accelerometer. Several external forms were developed, but they shared the sensing and control concept shown in Figure 1. The earlier toy used an ESP32-WROVER-E and heuristic response logic. All current machine-learning development and hardware measurements refer to a Seeed Studio XIAO ESP32-S3 operated at 240 MHz with 8 MB flash and 8 MB octal PSRAM.

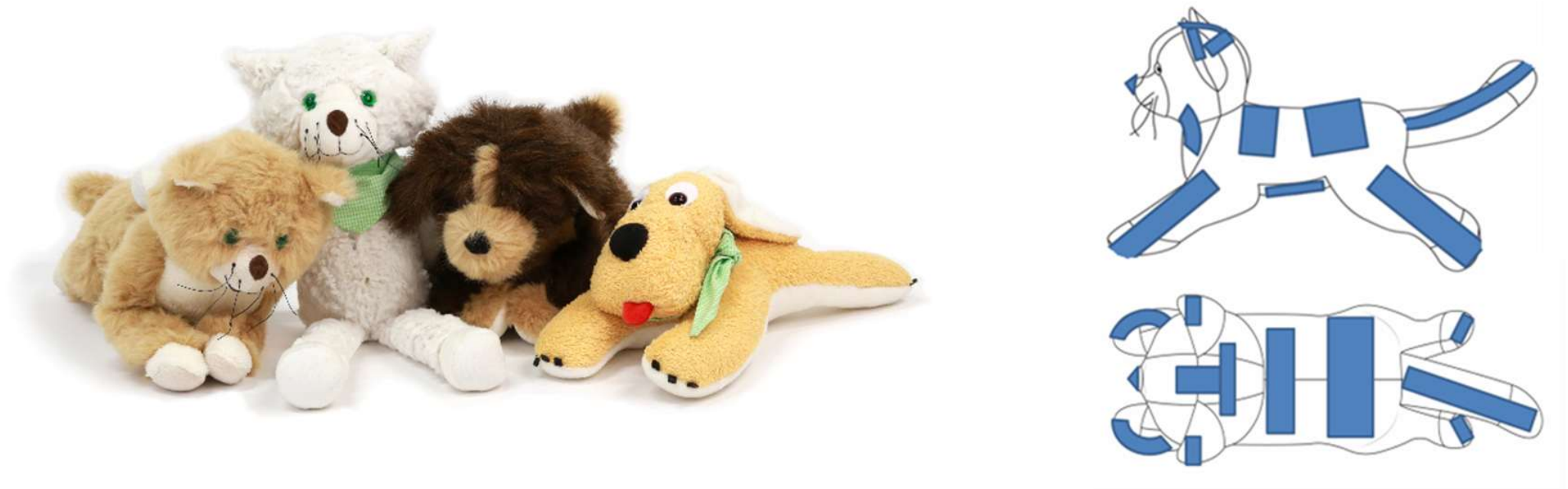

(a) (b)

**Figure 1. Soft plush companions and integrated capacitive-sensor placement: (a) developed companions; (b) representative sensor placement.**

Signals were sampled at 100 Hz. The communication frame contained ten capacitive sensor-derived values, the three accelerometer axes, a timestamp, and the inter-sample delay, as is presented in Table 1. Only 11 channels entered any classifier: the ten capacitive channels and one accelerometer magnitude $Acc_{magn} = \sqrt{x^2 + y^2 + z^2}$. Timestamp and delay were used for data handling and quality checks, not training. Five trailing zeros were added only inside the ESP-DL deployment tensor to align the 11 informative channels with 16-channel vector lanes; those values were padding rather than inputs.

**Table 1. Acquired signals and their use in the classifiers.**

| Stream field(s) | Signal | Classifier representation | Used for learning |
|---|---|---|---|
| 1–10 | Ten capacitive sensor-derived channels (1-cumulative; 2-cumulative extremities; 3-ears; 4-head; 5-neck; 6-proximity; 7-nose; 8-tail; 9-back; 10-belly) | Ten separate channels | Yes |
| 11–13 | Accelerometer x, y, z | One magnitude channel | Yes |
| 14 | MCU timestamp | Metadata / quality control | No |
| 15 | Delay since previous sample | Metadata / quality control | No |

The legacy heuristic mapped sensor amplitudes and durations to positive or negative feedback, expressed through eye colour and sound. Its response was sensitive to environmental conditions and individual interaction style, particularly for slow or spatially distributed contacts. The learned classifier was introduced to provide a richer gesture estimate; the heuristic remains a potentially useful fast path for unequivocal high-energy events.

### 3.2. Data Collection and Dataset Composition

Twenty-five participants contributed gesture recordings: 14 female and 11 male participants, including 8 adults, 9 teenagers, and 8 kindergarten-age children. Each participant performed 17 gestures three times, producing 51 planned human-interaction sequences per participant and 1,275 sequences in total. A further 51 at-rest recordings captured the toy without a human interaction in different orientations. Thus, 1,326 recordings were acquired before cleaning.

The data were genuinely collected for this project through supervised sessions rather than assembled from pre-existing datasets. Each participant session required approximately 20–30 min of direct recording time, corresponding to about 8–12 h of supervised acquisition, excluding recruitment, scheduling, permission, consent, and coordination with children and their guardians. The sample size was selected as a feasibility balance between demographic coverage, representativeness for model development, and the project's practical time constraints; it was not based on a formal power calculation.

**Table 2. Gesture definitions and operational class mapping.**

| 18-class label(s) | Description | 10-class label |
|---|---|---|
| At rest | Toy remains still in different orientations | At rest |
| Back stroke 1 / 2 | Back stroked on table / while held | Back stroke |
| Ear scratch 1 / 2 | Area between ears scratched on table / while held | Ear scratch |
| Head stroke 1 / 2 | Head stroked on table / while held | Head stroke |
| Hold in hands | Toy lifted and held | Hold in hands |
| Hold strong | Toy pressed firmly against the body | Hold strong |
| Neck scratch 1 / 2 | Neck scratched on table / while held | Neck scratch |
| Tail hit 1 / 2 | Energetic strike near tail on table / while held | Tail hit |
| Tail pull | Single energetic pull of the tail | Tail pull |
| Tail scratch 1 / 2 | Back near tail scratched | Tail scratch |
| Tail tap 1 / 2 | Back near tail tapped lightly | Tail scratch |

The 18-class taxonomy deliberately retained context variants and closely related gestures. It was used as a high-granularity stress test of compact temporal models. The operational 10-class taxonomy merged context variants that did not require different toy responses and combined tail scratching with tapping because both activated the same region and could share a response. Gesture variants that differed only in whether the toy was held in the participant's hands were also merged. This reformulation therefore reduced label granularity to the intended decision space.

The selected interactions follow established social-touch concepts used in human–robot interaction research [56], with additional emphasis on gestures that the legacy heuristic system found difficult.

*Recording procedure*

Data collection was automated using the *Gesture Recorder* tool, adapted from Jon Froehlich's Physical Computing repository [58]. The version adapted for the hardware of plush interactive toys can be found at repository [55] in folder "Processing". GUI (Graphical User Interface) of the tool is shown in Figure 2.

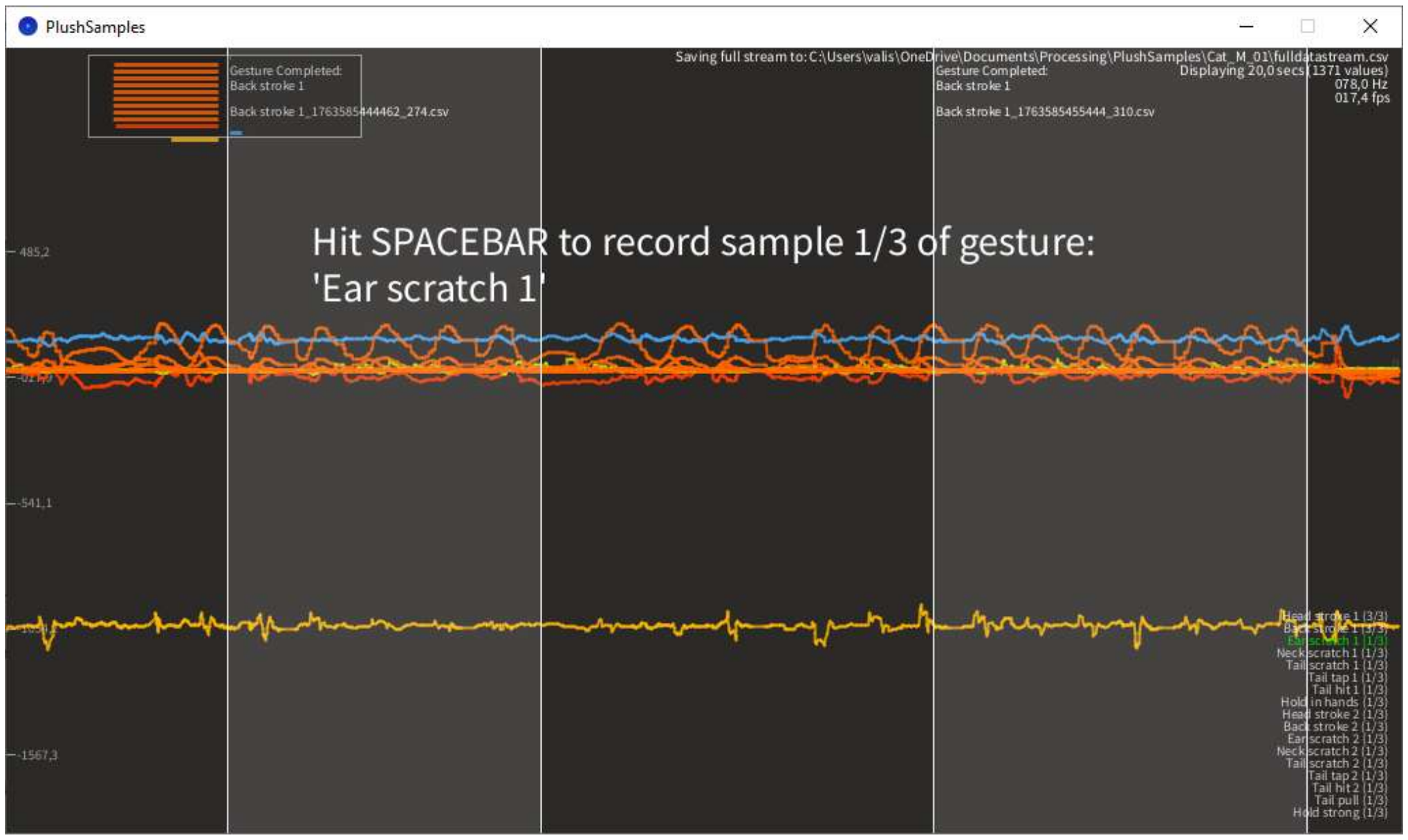

**Figure 2. Gesture Recorder interface used for supervised acquisition of labelled sensor sequences.**

Before a data recording session, the soft toy is paired with a computer via a Bluetooth link. The Bluetooth driver creates a serial port used for data exchange. The program establishes a link with the toy by opening this port, then it instructs the user which gesture needs to be recorded and records all sensor data. The program cycles through all gestures sequentially, generating labelled and timestamped CSV files for each sample.

*Preprocessing and data management*

Data preprocessing was implemented in MATLAB through a modular workflow and is available in the repository in folder "MATLAB" [55]:

1. **Anonymization and organization** – *Gesture Recorder* tool saves data obtained from each subject in a separate folder. Data from these folders containing samples of individual participants are copied into gesture-labelled directories using the script *a_move_labelled_data.m*, ensuring privacy and class consistency. A randomly generated ID is added to the files for anonymization and shuffling purposes. It should be noted that while providing privacy protection, this step makes it impossible to control data leakage during training, unless the necessary meta-data is preserved. For leakage-controlled, participant-independent cross-validation, a separate Python program was created that reconstructed the provenance link between each processed window and its source recording and participant.

2. **Quality control** – The *b_training_data_analysis.m* script enables one to compute summary statistics (mean, median, variance, and range) for each gesture class. Samples deviating by ±40% (can be set using *thresholdRange* variable) from mean duration are flagged as outliers and plotted automatically for user's rapid evaluation. Two such indications for the gesture "Hold strong" are shown in Figure 3. The user is responsible for manual review of these outliers.

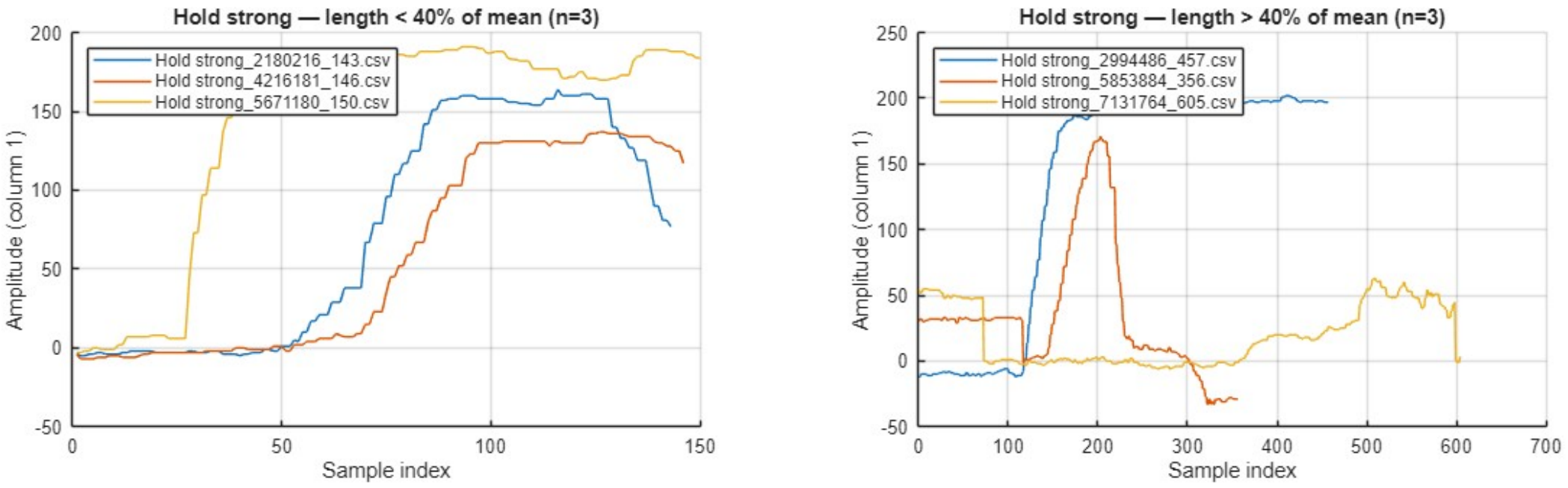

**Figure 3. Example output of the MATLAB quality-control routine for atypical recording duration or signal behaviour.**

3. **Truncation and alignment** – The script *c_training_data_truncation.m* provides an interactive tool for trimming the variable delay between the recording prompt and the onset of the intended gesture caused by human reaction time, see Figure 4. If the truncation is necessary, the user needs to click on the displayed plot of sample signal to indicate where the gesture starts. A vertical dotted line is displayed as a result. After pressing the Enter key, the next sample is plotted. This allows to streamline processing of the recorded samples, given that the total number of recorded samples usually is at least in hundreds and manual adjusting in a spreadsheet editor or by other means would be too time-consuming.

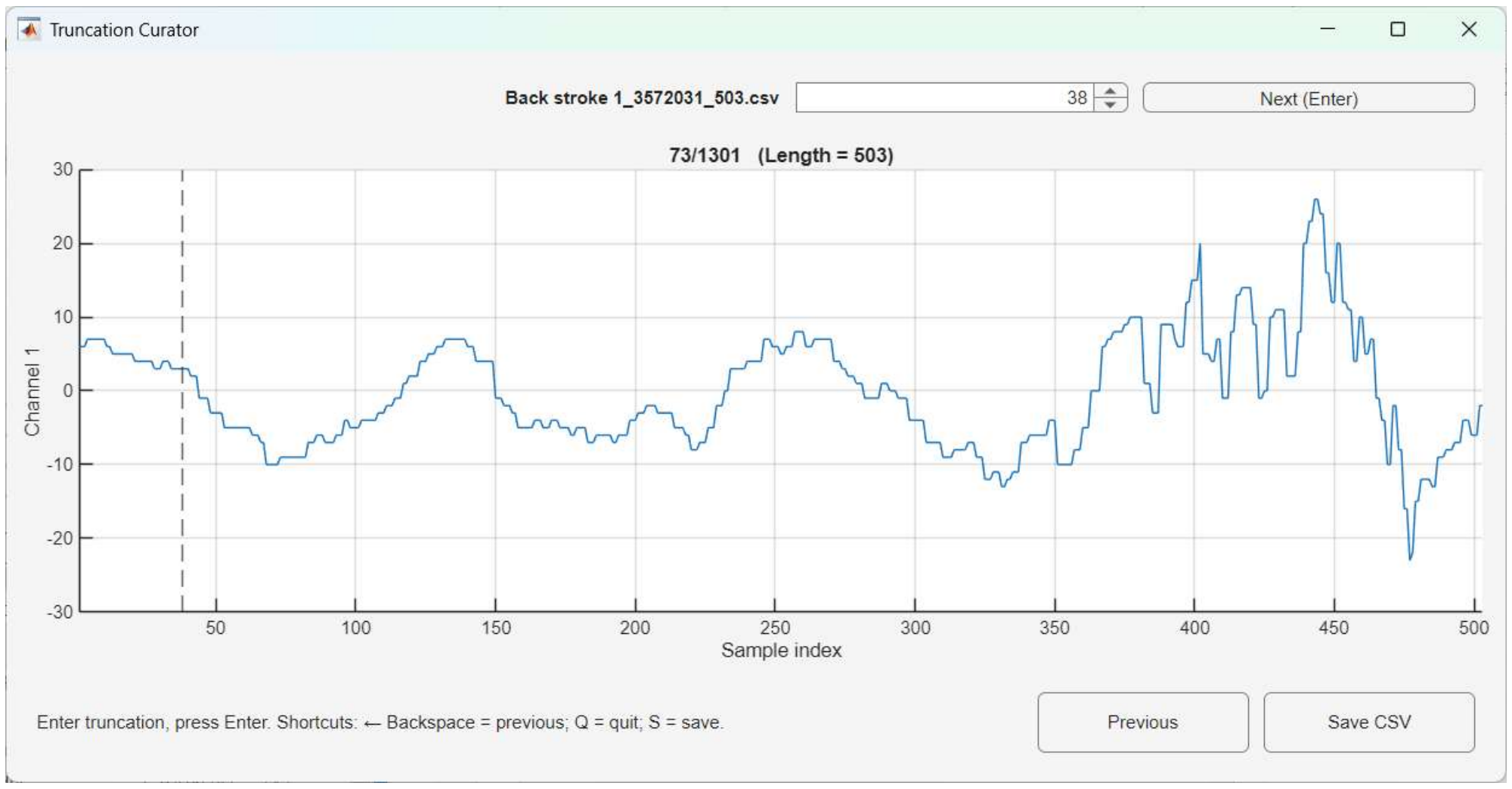

**Figure 4. MATLAB tool for interactively removing the initial reaction-time segment from a recording.**

4. **Windowing and augmentation** – Processed signals are segmented using a sliding window of 250 samples (2.5 s at 100 Hz) with a 50 sample step. Short sequences are looped to match window length rather than zero-padded, preserving physical signal properties. Figure 5 shows how a single signal with a total length of 700 can be divided into 10 distinct samples using such sliding window technique. The corresponding scripts can be found in *d_data_preparation.m*. This data augmentation approach makes the training and the resulting model more robust, since the training dataset contains samples with different temporal starting points.

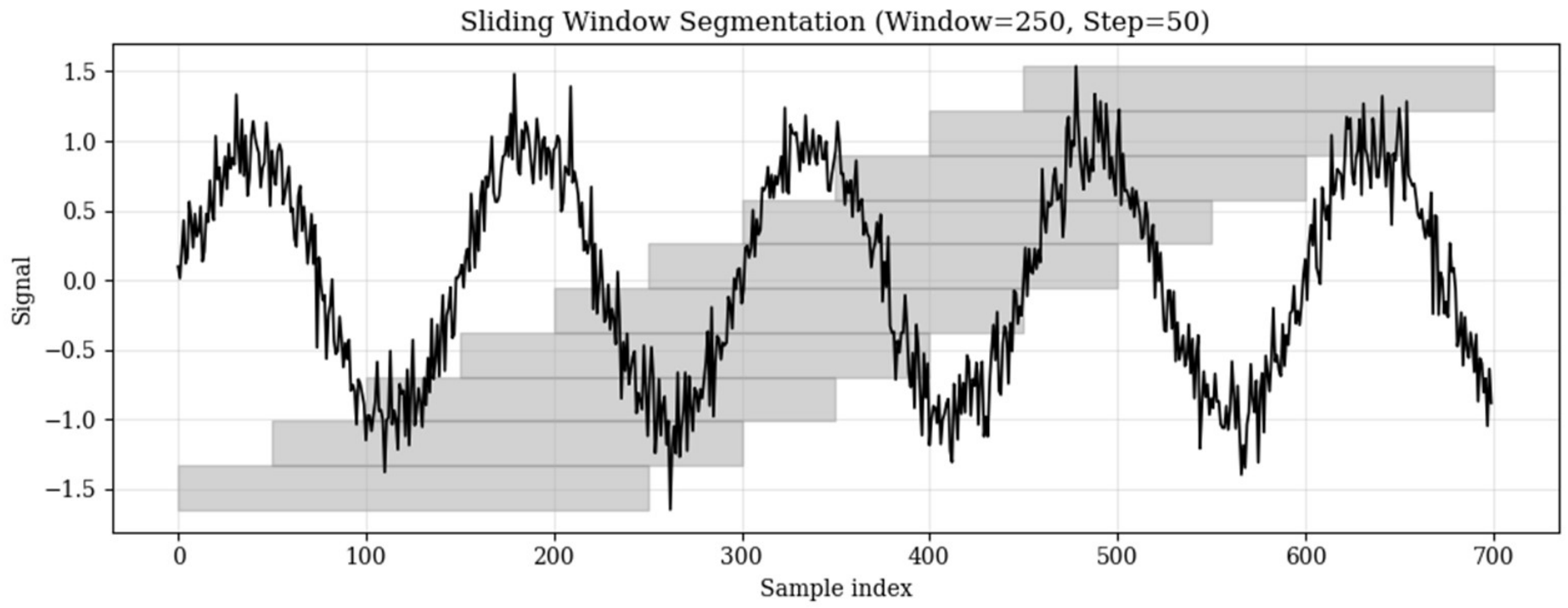

**Figure 5. Sliding-window generation using a 250-sample window and a 50-sample step.**

The windowing parameters are aligned with findings of other studies [30,31], who emphasize that the response should be fast enough. Moreover, this interval was confirmed by our initial tests, which have shown that 2.5 s was the least time that was necessary to distinguish between the gestures included in our training set. The 2.5-s window was a pragmatic interval observed during preliminary interaction trials to contain both brief and slowly executed gestures while retaining a responsive update interval. It was not selected by a formal window-length sensitivity study. A separate script *e_train.m* was developed for conducting such manual training sessions to test particular parameters or architectures. It enables one to train a particular architecture and check its accuracy on validation data.

### Ethical Considerations – Institutional Review Board Statement

The studies involving humans were approved by Ethics Committee of the Research Committee of Riga Technical University (protocol 04000-10.2.3-e/11, March 25, 2025). The studies were conducted in accordance with the local legislation and institutional requirements. Written informed consent for participation in this study was provided by the participants' legal guardians/next of kin.

### 3.3. MATLAB-Based Data Preparation and CNN Development Workflow

MATLAB was the core environment for the original model-development workflow. The pipeline anonymized and organized recordings, computed class-level summary statistics, flagged recordings for manual inspection, supported interactive trimming, generated fixed-length windows, trained individual networks, and automated architecture sweeps through *MATLAB Experiment Manager*. These components were important for this study: they transformed raw physical-toy signal recordings into auditable learning inputs and enabled a broad investigation of temporal receptive-field designs.

*Hyperparameter exploration*

The exploratory study evaluated 13 one- to four-layer CNN topologies under four channel-width settings, three epoch limits, and three batch sizes, giving 468 trials. Candidate networks used convolution, batch normalization, rectified linear activation, temporal aggregation, and a fully connected classification layer. Kernel widths and dilation factors were selected so that the theoretical receptive field approached the complete 250-sample window. To maintain direct correspondence between the manuscript, experimental outputs, and source code, the symbolic architecture identifiers used in the software are retained throughout the paper. For example, 14d1_39d2_41d4 denotes a three-layer configuration with kernel widths of 14, 39, and 41 and corresponding dilation factors of 1, 2, and 4, respectively. The number of filters in successive convolutional layers is normally specified as a tuple following this identifier.

This generated 468 trained networks, each evaluated on held-out test data. The primary evaluation metric was classification accuracy, supplemented by computational efficiency indicators such as total parameters. The results are presented in the following sections.

*Training setup*

Training used 70% of the dataset for learning, 15% for validation, and 15% for testing final accuracy. The Adam optimizer with learning rate 0.001 was used throughout. No early stopping criteria were set at this point to run the tests for the total set number of epochs. The rationale was to ensure uniformity in model training for further comparison. Early stopping criteria are introduced at a later stage during more detailed analysis of top performing models.

The experiment was executed on a workstation equipped with two NVIDIA RTX 2070 GPUs, completing training in approximately 48 hours. The final models were ranked according to accuracy and computational requirements to identify candidates suitable for microcontroller deployment.

Because applying 25-fold LOSO-CV to all 468 configurations would have required 11,700 training runs, subject-level separation was not imposed during this exploratory stage; the results were used only for coarse architecture screening, followed by accuracy–compute Pareto analysis, shortlisting, and participant-independent LOSO evaluation of the selected best-performing configurations.

### 3.4. Python Reimplementation and Participant-Aware Processing

A parallel Python workflow reimplemented the MATLAB stages with near one-to-one processing logic and several changes needed for strict evaluation and embedded export. The planned sequence comprises *00_explore.ipynb, 01_truncation.py, 02_preprocess.py, 03_train.py, 04_RT_test.py*, and *05_export.py*, which are included in the repository.

MATLAB provided the original environment for data acquisition, visual inspection, interactive truncation, window generation, architecture notation, and exploratory CNN evaluation. A parallel Python implementation extended this workflow to participant-aware confirmatory experiments and embedded deployment. It retained participant and recording identifiers, generated deterministic fold manifests, executed repeated training using CUDA acceleration on an NVIDIA GPU, and produced ONNX, ESP-DL, and C++ artifacts for direct integration and testing in the VS Code deployment environment.

Each of the 11 input channels was normalized independently using a z-score. For channel (c), the mean $\mu_c$ and sample standard deviation $\sigma_c$were calculated. Thus, the calculation pooled all temporal values from all training windows and produced one mean and one standard deviation for each input channel. Standard deviations below 1.0 in the original sensor units were replaced by 1.0 to prevent unstable scaling of nearly constant channels. Each value was then transformed as $Z = \frac{x-\mu_c}{\sigma_c}$ and clipped to [−3, 3].

The normalization parameters were calculated separately for every outer LOSO fold using its training partition only. The same fold-specific parameters were applied unchanged to the training, validation, and held-out test windows. Normalization was therefore performed per channel and per fold. For the linear SVM, the 66 extracted features received an additional standardization step whose means and standard deviations were likewise fitted on the training features only. The random forest used the channel-normalized features without this additional feature-scaling stage.

### 3.5. Evaluated Classification Models

The initial architecture search focused on dilated 1D CNNs as feature-free classifiers operating directly on multichannel sensor windows. For the confirmatory stage, the comparison was broadened to determine how the shortlisted CNNs performed against alternative compact neural architectures and traditional feature-based classifiers. The selected comparators – DS-CNN, TCN, compact Transformer, linear SVM, and constrained random forest – represented different approaches to temporal modelling, feature extraction, and computational efficiency while remaining relevant to ESP32-S3 deployment. This deliberately limited model set was evaluated using identical participant folds, input windows, class definitions, and performance criteria. See Table 3 for the summary.

#### 3.5.1. Shortlisted Dilated 1D CNNs

The four Pareto candidates comprised two wider three-layer networks, a compact 14d1_39d2_41d4 network with channel widths 8–12–16, and a smallest four-layer network with widths 4–6–8–8. The compact 8–12–16 network contained 13,650 parameters and required 3.352 million multiply–accumulate operations (MACs) per window. It was selected as the principal dilated-CNN comparator because it preserved the full-window receptive field at modest size.

#### 3.5.2. DS-CNN, TCN, and Compact Transformer

The DS-CNN began with a conventional strided convolution and then used three depthwise temporal convolutions followed by pointwise channel mixing; global average pooling fed a 10-class output. The compact TCN combined strided convolutions with residual dilated blocks. The compact Transformer reduced the 250-sample sequence to 25 tokens using a strided patch embedding and applied one pre-normalized two-head attention block with model width 32 and feed-forward width 64.

#### 3.5.3. Linear SVM and Constrained Random Forest

Traditional models used 66 predefined time-domain features: mean, population standard deviation, minimum, maximum, root-mean-square value, and mean absolute first difference for each of the 11 normalized channels. The one-versus-rest linear SVM selected C from {0.01, 0.1, 1, 10}; its feature scaler was fitted on the training partition only. The random forest selected 16, 32, or 64 trees, maximum depth 4, 6, or 8, and minimum leaf size 1, 3, or 5, with Gini impurity, square-root feature subsampling, and balanced class weights. Validation macro-F1 determined hyperparameters inside each outer fold, with deterministic tie breaking.

**Table 3. Models in the final 10-class comparison.**

| Model | Input | Main configuration | Parameters | MAC/window |
|---|---|---|---|---|
| Dilated CNN | 250 × 11 raw window | 14d1_39d2_41d4 (8, 12, 16) | 13,650 | 3,352,160 |
| DS-CNN | 250 × 11 raw window | Strided stem + 3 depthwise-separable blocks | 13,602 | 912,280 |
| TCN | 250 × 11 raw window | Strided stem + residual dilated blocks | 30,090 | 1,236,128 |
| Compact Transformer | 250 × 11 raw window | 25 tokens; d=32; 2 heads; 1 block | 14,026 | 373,120 |
| Linear SVM | 66 handcrafted features | One-versus-rest linear scores | — | — |
| Random forest | 66 handcrafted features | Bounded ensemble of shallow trees | — | — |

### 3.6. Experimental Design

#### 3.6.1. Exploratory 468-Run CNN Search

The exploratory CNN search was implemented in MATLAB Experiment Manager using the developed *PlushCNNExperiment.mlx* experiment definition. The script parsed symbolic architecture identifiers, constructed the corresponding layer graphs, assigned the selected training parameters, and executed and logged the trials automatically. Thirteen materially different temporal-convolution configurations were evaluated: 250, 126d2, 100_151, 83d1_84d2, 1d1_126d2, 70_90_91, 14d1_39d2_41d4, 28d1_32d2_41d4, 20d2_16d4_20d8, 19d1_24d3_19d9, 1d1_84d3_1d1, 1d1_27d3_20d9, and 19d1_19d2_19d4_17d8. These configurations covered one to four convolutional layers, undilated and dilated convolutions, several dilation schedules, and bottleneck structures beginning with a single filter.

The kernel widths and dilation factors were selected so that the theoretical receptive field of each topology approximately matched the 250-sample input window, ranging from 249 to 255 samples. This allowed each candidate to represent temporal relationships spanning almost the complete 2.5-second observation interval while testing different distributions of local and long-range temporal processing. The architecture symbolic identifiers follow the notation used directly in the software; for example, 14d1_39d2_41d4 denotes kernel widths of 14, 39, and 41 with dilation factors of 1, 2, and 4.

Each of the 13 topologies was evaluated with four filter-count configurations—(4, 6, 8, 8), (8, 12, 16, 16), (16, 24, 32, 32), and (32, 48, 64, 64)—three maximum epoch counts (100, 250, and 1000), and three mini-batch sizes (64, 256, and 3936). Values corresponding to nonexistent layers were ignored for architectures containing fewer than four convolutional layers. The resulting Cartesian product comprised 13 × 4 × 3 × 3 = 468 training trials. All trials used the Adam optimizer with an initial learning rate of 0.001. Early stopping was not applied at this exploratory stage, allowing configurations with different epoch limits to be compared under consistently specified training durations.

For this initial search, windows were randomly assigned to training, validation, and held-out subsets in proportions of 70%, 15%, and 15%, respectively. MATLAB Experiment Manager recorded the accuracy and loss histories and retained the trained-network metadata for subsequent comparison. Because overlapping windows originating from the same participant or recording could occur in different subsets, these results were used only for broad architecture screening and computational-complexity analysis.

Candidate selection considered held-out accuracy together with parameter count and MAC requirements, producing a set of non-dominated dilated-CNN configurations for the subsequent participant-independent LOSO evaluation.

#### 3.6.2. Eighteen-Class LOSO Evaluation

Each shortlisted CNN was retrained in 25 LOSO folds on the original 18-class taxonomy. In each fold, one participant supplied the gesture test set, four different participants supplied validation data, and the remaining 20 supplied training data. No participant or derived window crossed these partitions. At-rest recordings were subjectless and were split separately into 80% training, 10% validation, and 10% secondary test windows; primary LOSO metrics excluded at-rest samples.

All partitions were normalized using the fold-specific procedure defined in Section 3.4.

#### 3.6.3. Operational 10-Class Reformulation

The same fold manifests and preprocessing rules were then applied to the response-equivalent 10-class taxonomy. This stage tested whether a task aligned with the toy's intended decision policy improved generalization without discarding participants or windows.

#### 3.6.4. Six-Model 25-Fold Comparison

The compact dilated CNN, DS-CNN, TCN, compact Transformer, linear SVM, and constrained random forest were compared on identical 25-fold splits and the same 4,268 held-out non-rest windows. Neural and traditional hyperparameters were selected using validation participants only. Final held-out predictions were generated once and stored as paired out-of-fold (OOF) predictions.

#### 3.6.5. Neural Training Hyperparameters

Neural models used AdamW with initial learning rate 0.001, weight decay 0.0001, batch size 128, inverse-frequency class weighting, weighted cross-entropy, and label smoothing 0.02. A cosine schedule reduced the learning rate to 0.00001 over at most 80 epochs. Early stopping monitored non-rest validation macro-F1 with patience 10 and minimum improvement 0.002. Training ran with PyTorch 2.14 on a CUDA-compatible NVIDIA GPU.

### 3.7. Evaluation Metrics and Statistical Analysis

Macro-averaged F1 was the primary participant-level endpoint because class support was unequal and several gestures had few windows for some participants. Accuracy and balanced accuracy were reported for continuity. Precision, recall, F1, and support were computed per class. Mean and sample standard deviation were calculated across the 25 held-out participants. The 95% confidence intervals were calculated across the 25 participant-level fold estimates using Student's $t$ distribution with 24 degrees of freedom. Both mean LOSO-fold accuracy and pooled OOF accuracy were reported. Mean fold accuracy assigns equal weight to each of the 25 held-out participants, whereas pooled OOF accuracy aggregates all 4,268 held-out predictions and therefore weights participants according to their number of windows.

Paired fold-level macro-F1 values were compared with a Friedman test across all six models. Following a significant omnibus result, planned two-sided Wilcoxon signed-rank comparisons against the compact dilated CNN were corrected by the Holm procedure. Hardware outcomes included serialized model bytes, model-context memory where meaningfully measurable, model-only inference, preprocessing or feature-extraction time, complete classifier time, and host–device label agreement.

Deployment cost is reported as parameter count, MACs per window, RAM footprint, and inference time measured on the ESP32-S3 edge device. We use this formula for the calculation of MAC of a CNN model:

$$MAC = 250 \sum_{l} ( K_l C_{l-1} C_l ) + 18 C_{last}$$

### 3.8. PC-Based Real-Time Simulation of the Trained Models

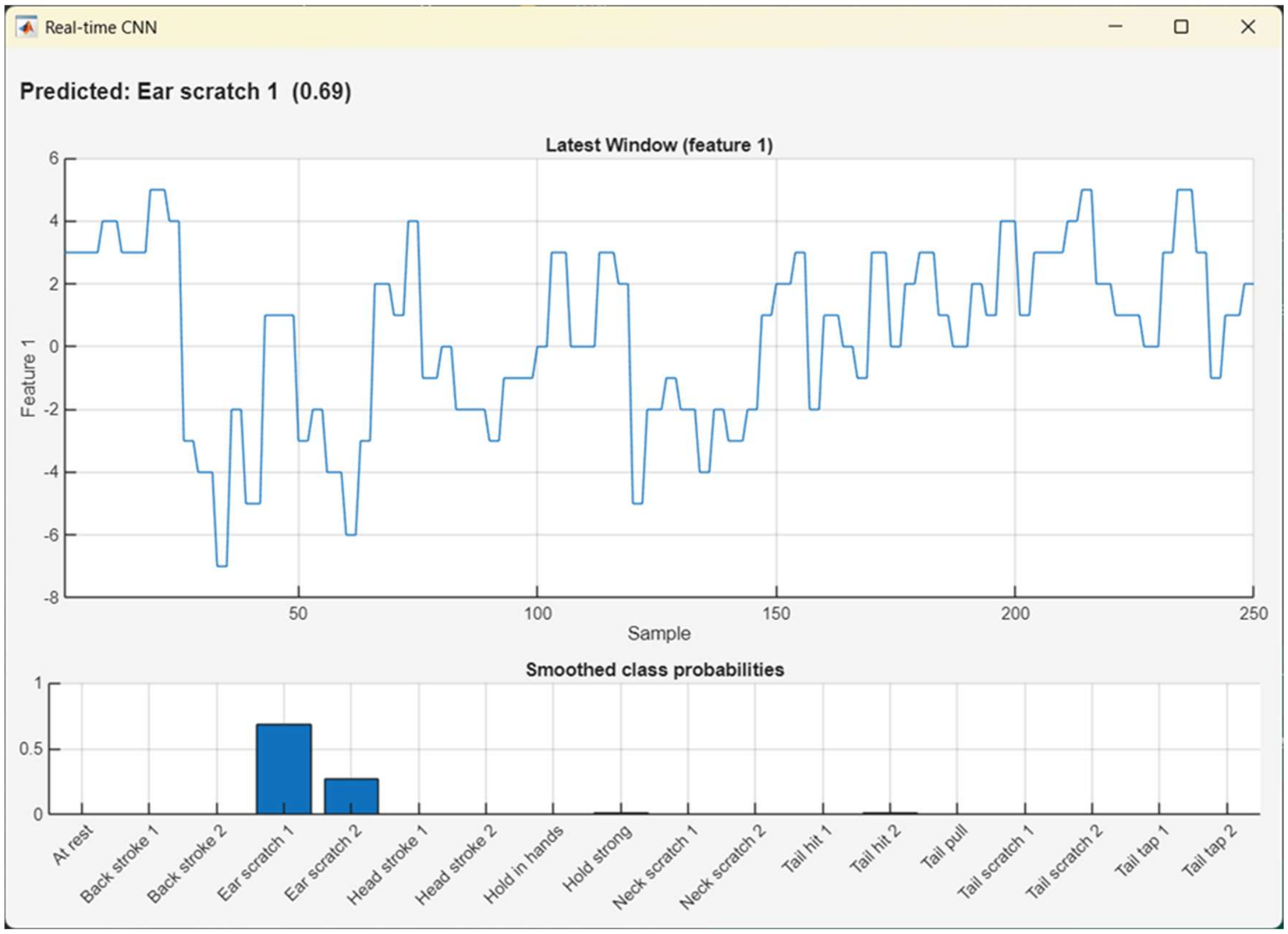

**Figure 6. MATLAB interface used for formative inspection of live sensor streams and model probabilities.**

Selected models were connected to the live sensor stream through a real-time simulator implemented in MATLAB (*f_realtime_classifier.m*). GUI of this classifier is shown in Figure 6. This tool was used to inspect prediction behaviour while interacting with the physical toy and to compare it informally with legacy heuristic responses. Subtle contacts were often recognized by the learned model when threshold rules reacted slowly or not at all; abrupt energetic events remained easy for the heuristic. Gesture onset, participant behaviour, and response time were not instrumented, so this activity is reported as formative qualitative observation rather than a quantitative real-time experiment.

### 3.9. ESP32-S3 Export, Parity Verification, and Hardware Benchmarking

Neural models were exported through static ONNX graphs and quantized with symmetric W8A8 (*int8* weights, *int8* activations). Neural-model inference was implemented using ESP-DL 3.3.10 within the ESP-IDF 6.0 development framework. Traditional models were emitted as static FP32 C++ arrays. Before a timing result was accepted, the exported host implementation and the ESP32-S3 were evaluated on the same deterministic fixture of 100 real windows, ten per operational class. This fixture tested numerical portability and timing.

Published ESP-DSP benchmark information [64] and work on optimized low-precision microcontroller kernels [65] motivated the target choice, but no latency value in this study was inferred from those sources: every reported classifier time was measured directly on the ESP32-S3.

The Seeed Studio XIAO ESP32-S3 operated at 240 MHz. Three untimed warm-up executions were performed before measurement. Model-execution latency was measured separately for every window. Where supported by the benchmark harness, preprocessing and the complete path from a raw window through preprocessing to model execution were timed in separate passes. For the SVM and random forest, the measured stages comprised channel normalization, extraction of the 66 time-domain features, model prediction, and the complete classification path. Serial output and other toy-control operations were excluded from the timed intervals.

## 4. Results and Analysis

### 4.1. Exploratory 1D CNN Architecture Search

The 468-run MATLAB screen showed that multiple dilated architectures achieved useful random-window accuracy over a wide range of model widths. Larger networks did not guarantee proportionate improvement, and dilation provided complete-window context with fewer layers than undilated alternatives. Because the exploratory split mixed participant-derived windows, these values served only to reduce the design space. A subset of the 12 highest-performing configurations (see Table 4) was therefore examined jointly in terms of accuracy and computational demand, from which four non-dominated accuracy–compute candidates were selected for participant-independent testing.

**Table 4. Top CNN performers in all the model size classes**

| Symbolic 1D CNN architecture | Number of filters per layer L1 L2 L3 [L4] | Number of parameters | Maximum accuracy on test data | Average accuracy across hyperparameters | MAC |
|---|---|---|---|---|---|
| 14d1_39d2_41d4 | 32 48 64 | 192k | 79.0% | 66.7% | 47.70 M |
| 19d1_24d3_19d9 | 32 48 64 | 103k | 77.1% | 69.4% | 25.48 M |
| 28d1_32d2_41d4 | 32 48 64 | 186k | 77.1% | 68.2% | 46.24 M |
| 14d1_39d2_41d4 | 8 12 16 | 13.2k | 75.0% | 65.7% | 3.21 M |
| 19d1_24d3_19d9 | 16 24 32 | 27.9k | 74.1% | 65.5% | 6.79 M |
| 28d1_32d2_41d4 | 8 12 16 | 13.8k | 73.6% | 68.2% | 3.35 M |
| 19d1_19d2_19d4_17d8 | 4 6 8 8 | 3.4k | 72.7% | 59.1% | 0.82 M |
| 14d1_39d2_41d4 | 16 24 32 | 49.7k | 72.4% | 67.2% | 12.23 M |
| 28d1_32d2_41d4 | 16 24 32 | 49.5k | 72.3% | 67.4% | 12.18 M |
| 19d1_24d3_19d9 | 8 12 16 | 8k | 71.7% | 63.5% | 1.91 M |
| 28d1_32d2_41d4 | 4 6 8 | 4.1k | 69.1% | 57.0% | 0.99 M |
| 70_90_91 | 4 6 8 | 9.8k | 68.7% | 56.7% | 2.40 M |

Further analysis results in the identification of four accuracy–compute Pareto candidates for subject-independent LOSO testing, see Figure 7.

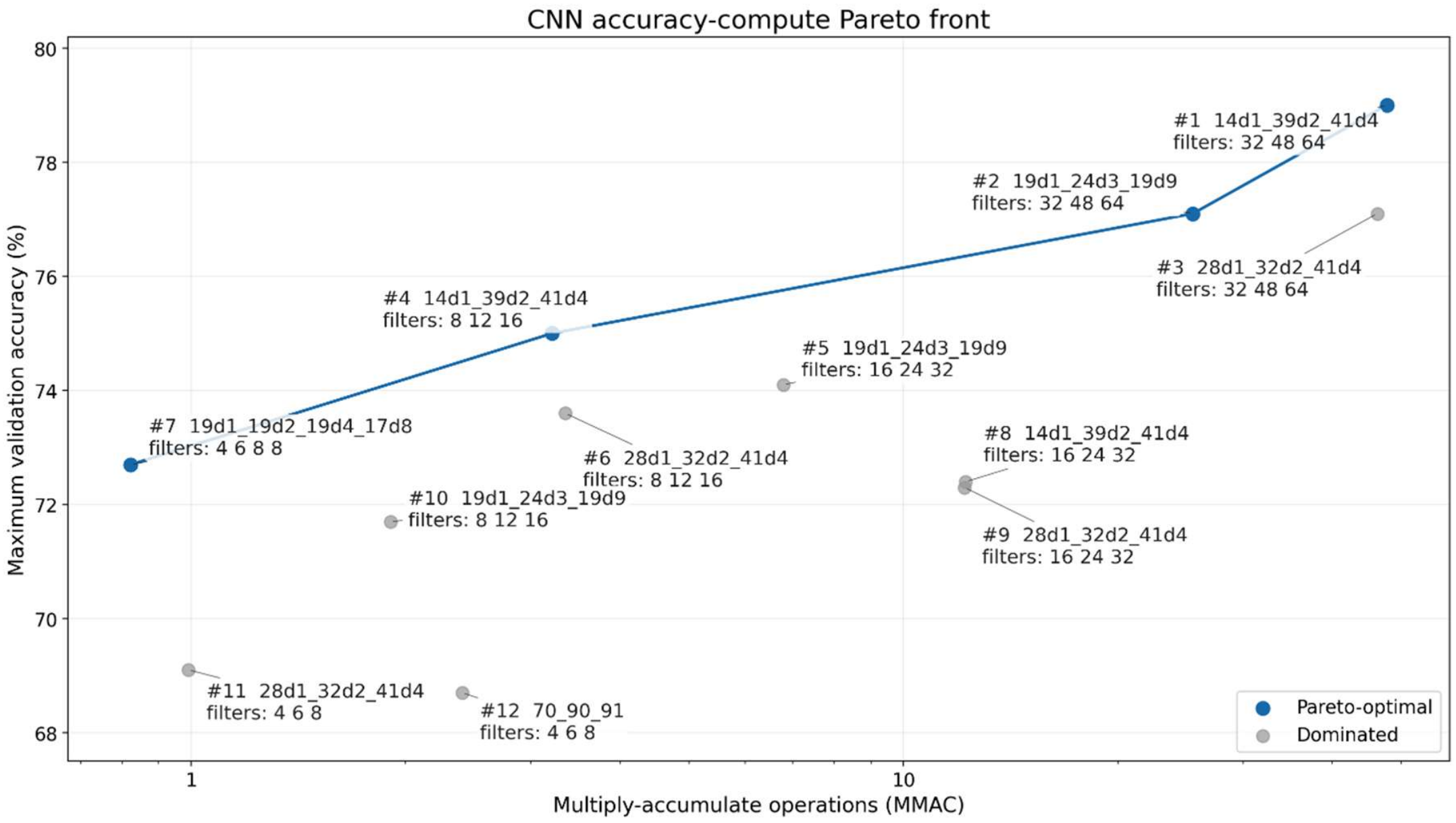

**Figure 7. Accuracy–compute Pareto analysis from the exploratory CNN screen. Highlighted candidates proceeded to LOSO evaluation; exploratory accuracy is not an unseen-subject estimate.**

### 4.2. Pareto Selection and 18-Class Subject-Independent Performance

The deliberately difficult 18-class task exposed the cost of distinguishing response-equivalent context variants and fine tail actions. The strongest mean fold accuracy was 71.48% for the 19d1_24d3_19d9 network, and its mean macro-F1 was 68.50% (see Table 5). The compact 13,650-parameter network achieved 68.62% accuracy and 64.53% macro-F1. Thus, compression reduced performance modestly, while the smallest model showed a clearer decline.

**Table 5. Twenty-five-fold LOSO results for shortlisted CNNs on the 18- and 10-class tasks.**

| Architecture | Filters | MAC | 18-class accuracy | 18-class macro-F1 | 10-class accuracy | 10-class macro-F1 | Inference time (average) |
|---|---|---|---|---|---|---|---|
| 14d1_39d2_41d4 | 32–48–64 | 48.26 M | 71.19 ± 13.43% | 66.80 ± 13.64% | 87.59 ± 10.40% | 84.47 ± 10.44% | 637.9 ms |
| 19d1_24d3_19d9 | 32–48–64 | 26.24 M | 71.48 ± 14.03% | 68.50 ± 14.21% | 86.39 ± 10.84% | 83.01 ± 10.97% | 196.9 ms |
| 14d1_39d2_41d4 | 8–12–16 | 3.35 M | 68.62 ± 12.94% | 64.53 ± 11.87% | 85.42 ± 11.79% | 81.96 ± 13.17% | 18.6 ms |
| 19d1_19d2_19d4_17d8 | 4–6–8–8 | 0.92 M | 60.80 ± 11.39% | 57.82 ± 11.02% | 80.64 ± 12.82% | 76.64 ± 14.31% | 14.6 ms |

Data in table 5 shows that the two larger models have inference times that make them practically unsuitable for hardware deployment. On the other hand, the accuracy of the third model is very similar to that of the top two performers. The difference from the second model is just around 1%, while the computation time is an order of magnitude smaller. The fourth model has even lower inference time, but its accuracy is substantially lower.

Due to practical reasons of our actual hardware implementation and initial understanding that a suitable ML model would need to have inference time of under 20 ms in order to implement it in the current version of hardware, model with symbolic configuration 14d1_39d2_41d4 and 8-12-16 filters was chosen for further deeper analysis and comparison to alternative models.

Figure 8 presents the confusion matrix for the 14d1_39d2_41d4 model with the 8-12-16 filter configuration. Because the "at rest" recordings were not associated with individual participants, this class is shown on only one axis to avoid implying that it was evaluated using the participant-based LOSO protocol. In addition to occasional isolated errors, the matrix reveals several clusters of systematically confused classes. For example, Head stroke 1 and Head stroke 2 differ only in whether the interactive toy was held in the participant's hands or placed on a table; the same distinction applies to Neck scratch 1 and Neck scratch 2. Another confusion cluster comprises Tail scratch 1/2 and Tail tap 1/2. These gestures produce similar sensor-response patterns, making their spatial and temporal separation particularly challenging.

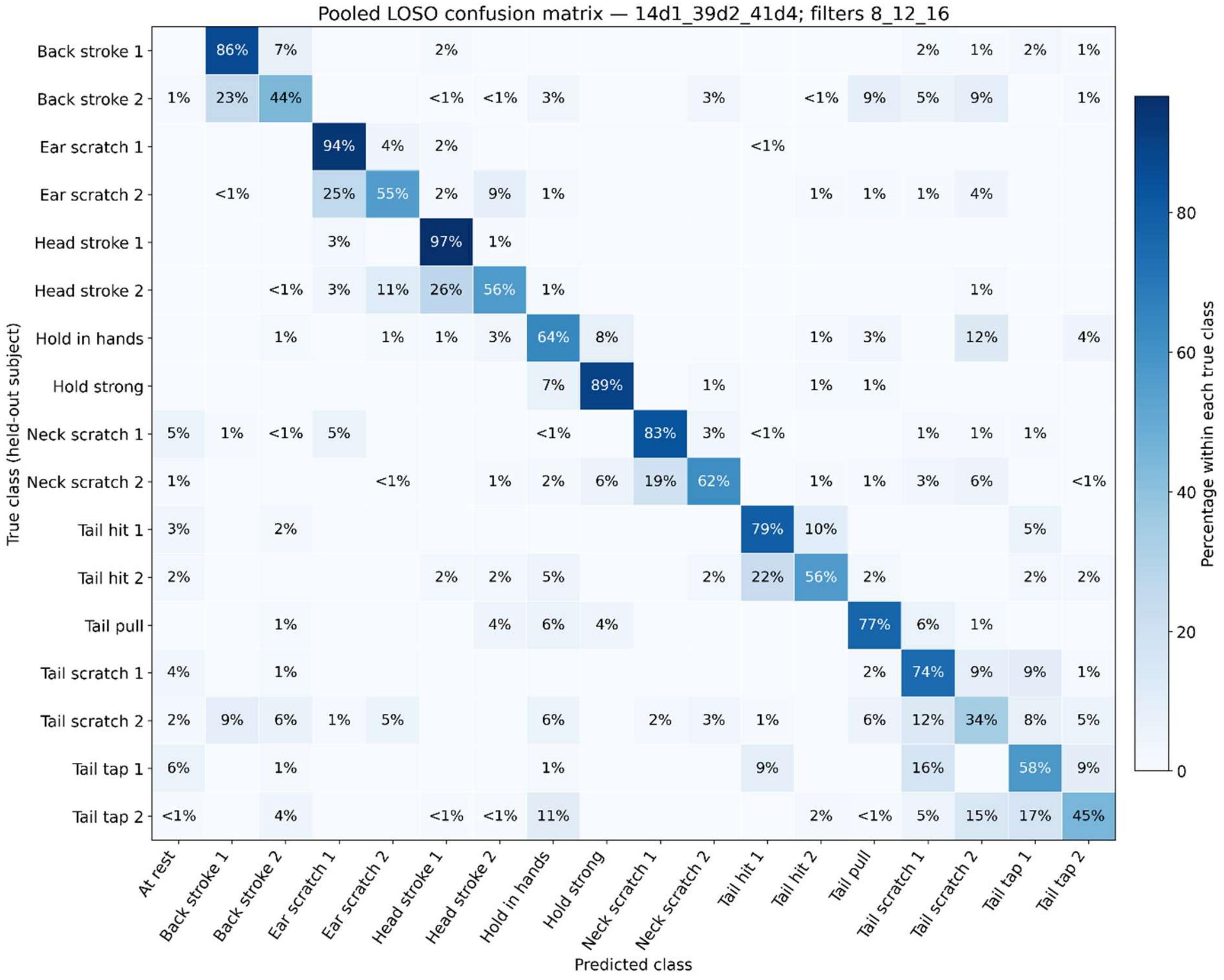

**Figure 8. Row-normalized pooled OOF confusion matrix for an 18-class CNN.**

Importantly, these systematic confusions occurred primarily between gestures assigned to the same emotional category, whether positive, negative, or neutral. They therefore reduced fine-grained gesture-classification performance but generally did not alter the emotional interpretation used by the toy's response system. We hypothesized that merging gesture classes with overlapping spatial and temporal activation patterns would simplify the classification task and improve quantitative performance by removing distinctions that were neither consistently expressed by participants nor operationally necessary. Because part of the observed overlap may reflect individual differences in how similar instructions were interpreted and performed, such merging was not expected to diminish the practical utility of the deployed classifier.

A further confusion cluster was observed between *Ear scratch 1/2* and *Head stroke 1/2*. The corresponding instructions differed mainly in whether participants stroked the toy's head with the palm or lightly scratched the nearby region with their fingers. Although these actions produced similar sensor-activation patterns, they were retained as separate classes because distinguishing them provided a demanding test of the model's ability to identify subtle differences between closely related gestures.

Information on how 18 classes are merged into 10 classes can be found in Table 2.

### 4.3. Operational 10-Class Reformulation

Merging response-equivalent labels increased mean macro-F1 for every shortlisted CNN by approximately 15–19 percentage points. The wider 14d1_39d2_41d4 model achieved the highest CNN result, with 87.59% mean accuracy and 84.47% mean macro-F1. The compact version retained 81.96% macro-F1 at less than one tenth of the wider model's MAC count. The gain supports the interpretation that the 18-class task encoded distinctions of limited operational value rather than revealing a general failure of temporal modelling.

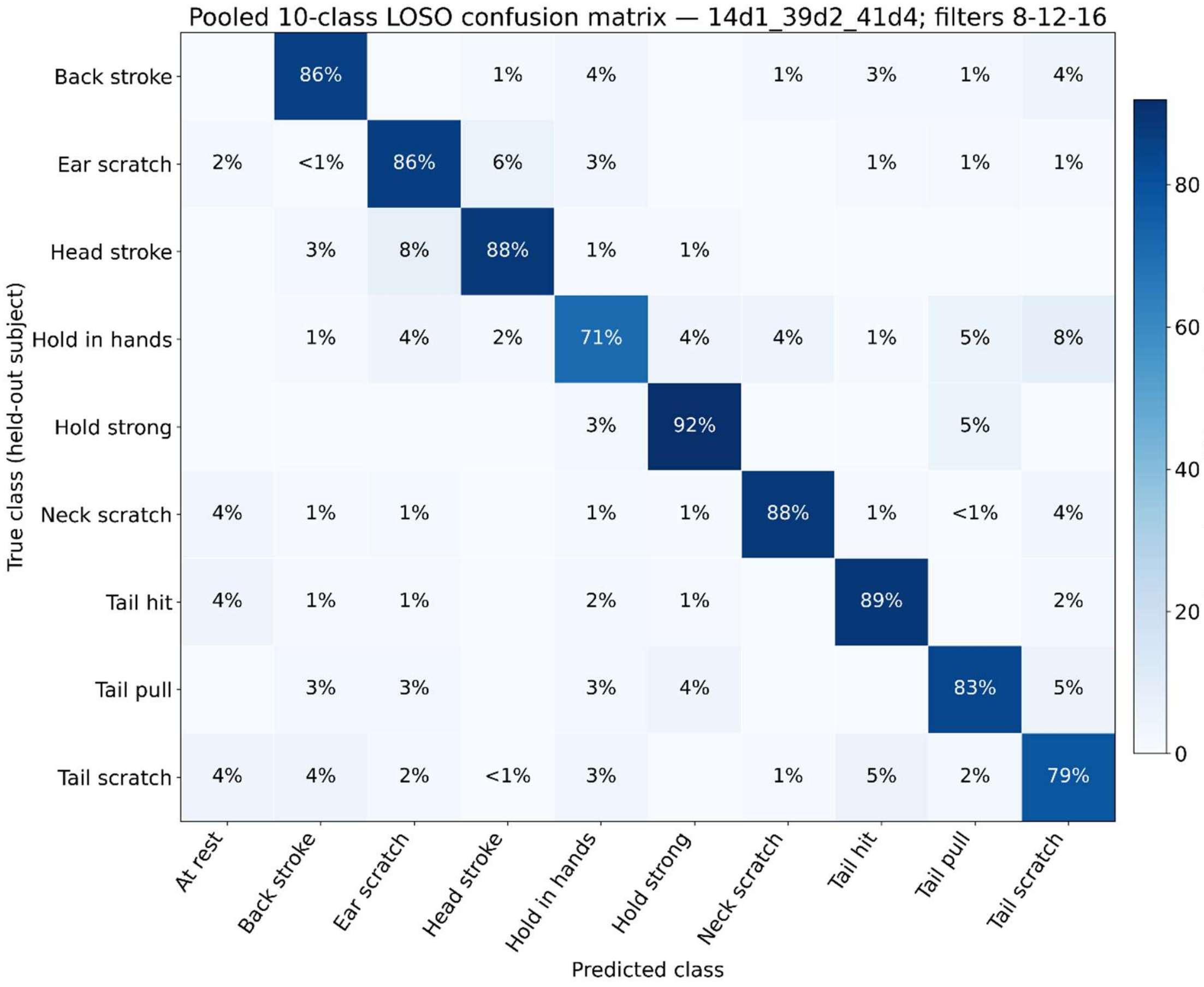

**Figure 9. Row-normalized pooled OOF confusion matrix for a 10-class CNN.**

Table 5 summarizes model performance on the original 18-class task and the reduced-granularity 10-class task. Figure 9 presents the confusion matrix for a 1D CNN evaluated on the 10-class problem.

As can be seen, merging similar classes had a positive effect on models' ability to find boundaries between distinct classes and it was not as confused by essentially the same gestures, which differed only by the position of the plush toy. Remarkable is the ability of the model to distinguish between classes "Ear scratch" and "Head stroke", which activate essentially the same sensors, but differ only by the fine movement of the palm or the fingers.

### 4.4. Subject-Independent Comparison of Model Families

Across the six-model comparison, the linear SVM achieved the highest mean fold accuracy (91.05%) and macro-F1 (87.98%). The random forest followed at 89.94% accuracy and 86.57% macro-F1. Among raw-window neural models, DS-CNN was strongest at 87.26% accuracy and 84.50% macro-F1; TCN achieved 83.33% macro-F1, the compact dilated CNN 81.96%, and the compact Transformer 79.44%. The exact pooled SVM accuracy was 3,841/4,268 = 89.995%; it is lower than the equally weighted mean fold accuracy because participants contributed different numbers of windows. The results are summarized in Table 6.

**Table 6. Six-model comparison on identical 25-fold, 10-class LOSO splits.**

| Model | Parameters | MAC/window | Fold accuracy | Fold macro-F1 | 95% CI of macro-F1 | Pooled OOF accuracy |
|---|---|---|---|---|---|---|
| Dilated CNN 14d1_39d2_41d4 (8–12–16) | 13,650 | 3.352 M | 85.42 ± 11.79% | 81.96 ± 13.17% | 76.53–87.40% | 84.33% |
| DS-CNN | 13,602 | 0.912 M | 87.26 ± 9.81% | 84.50 ± 9.65% | 80.52–88.49% | 86.29% |
| TCN | 30,090 | 1.236 M | 86.12 ± 10.46% | 83.33 ± 10.09% | 79.17–87.50% | 85.07% |
| Compact Transformer | 14,026 | 0.373 M | 84.13 ± 11.01% | 79.44 ± 10.65% | 75.05–83.84% | 83.22% |
| Linear SVM | — | — | 91.05 ± 9.25% | 87.98 ± 11.40% | 83.27–92.69% | 90.00% |
| Random forest | — | — | 89.94 ± 9.19% | 86.57 ± 9.18% | 82.78–90.36% | 89.18% |

Mean performance alone does not show how consistently a model generalizes across individual participants. Because every model was evaluated using the same 25 held-out participants, the resulting fold-level macro-F1 values constituted matched observations. Figure 10 presents their distributions and shows both the differences in central performance and the substantial between-participant variability. Boxes represent the interquartile range, horizontal lines indicate medians, green triangles indicate means, whiskers extend to 1.5 times the interquartile range, and the isolated circle denotes an outlier. The feature-based classifiers have higher central values, although their distributions overlap with that of the neural models.

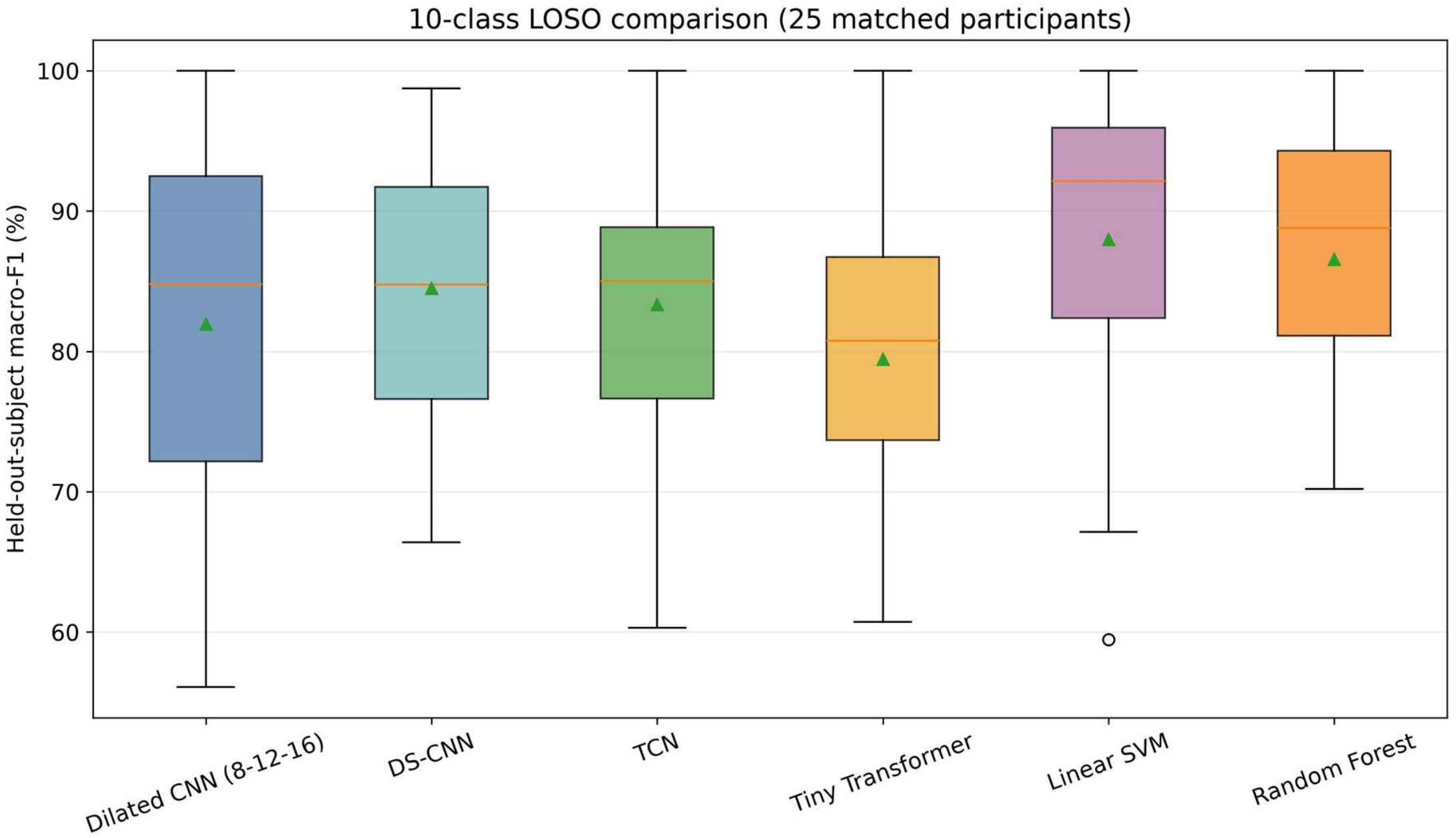

**Figure 10. Distribution of held-out-participant macro-F1 across the six models.**

Substantial between-participant variation was observed for all six models. The linear SVM and random forest achieved the highest central values, whereas the distributions of the four neural models overlapped considerably. Nevertheless, the compact dilated CNN was not uniformly outperformed by the feature-based classifiers. It achieved a higher macro-F1 than the linear SVM for five held-out participants and a higher macro-F1 than the random forest for nine participants, with one additional tie in each comparison.

The Friedman test identified a significant overall model effect, $\chi(5) = 31.41, p = 7.76 \times 10^{-6}$. Subsequent two-sided Wilcoxon signed-rank comparisons with the compact dilated CNN were adjusted using the Holm procedure. The differences were significant for the linear SVM (adjusted $p$ = 0.0027) and random forest (adjusted $p$ = 0.0477), but not for the DS-CNN ($p$ = 0.3405), TCN ($p$ = 0.5303), or compact Transformer ($p$ = 0.5303). These results support an overall advantage of the tested feature-based classifiers over the compact dilated CNN, while the numerical ordering among the tested neural architectures should not be interpreted as evidence that one neural family universally outperforms the others.

### 4.5. Per-Class Performance and Confusion Analysis of Linear SVM

The linear SVM was selected for detailed per-class analysis because it achieved the highest mean LOSO accuracy and macro-F1 among the six evaluated models. It also produced the smallest serialized classifier and the shortest end-to-end inference time on the ESP32-S3. Examining its class-level predictions therefore provides insight into the remaining recognition errors of the strongest predictive and deployment candidate identified in this study.

The linear SVM exceeded 0.90 F1 for back stroke, ear scratch, head stroke, neck scratch, and tail scratch. The lower-support classes remained harder: hold in hands reached 0.696 F1, tail hit 0.780, and tail pull 0.805. These errors are consistent with unequal class support and partial similarity between spatial contact patterns. The complete class-wise values for the principal model are reported in Table 7.

**Table 7. Pooled OOF per-class performance of the linear SVM on held-out human-gesture windows.**

| Class | Precision | Recall | F1 | Support |
|---|---|---|---|---|
| Back stroke | 0.935 | 0.917 | 0.925 | 623 |
| Ear scratch | 0.915 | 0.908 | 0.911 | 737 |
| Head stroke | 0.932 | 0.915 | 0.924 | 674 |
| Hold in hands | 0.609 | 0.813 | 0.696 | 144 |
| Hold strong | 0.840 | 0.919 | 0.877 | 74 |
| Neck scratch | 0.974 | 0.907 | 0.939 | 708 |
| Tail hit | 0.714 | 0.859 | 0.780 | 99 |
| Tail pull | 0.747 | 0.872 | 0.805 | 78 |
| Tail scratch | 0.937 | 0.888 | 0.912 | 1,131 |

To construct the confusion matrix, the out-of-fold predictions from all 25 held-out participants were pooled and normalized within each true gesture class. Figure 11 consequently represents predictions for windows that were never used for training. Rows represent true gesture classes from held-out participants, whereas columns represent the predicted operational classes. The at-rest category appears only as a prediction column because it was not included in the participant-level LOSO test sets.

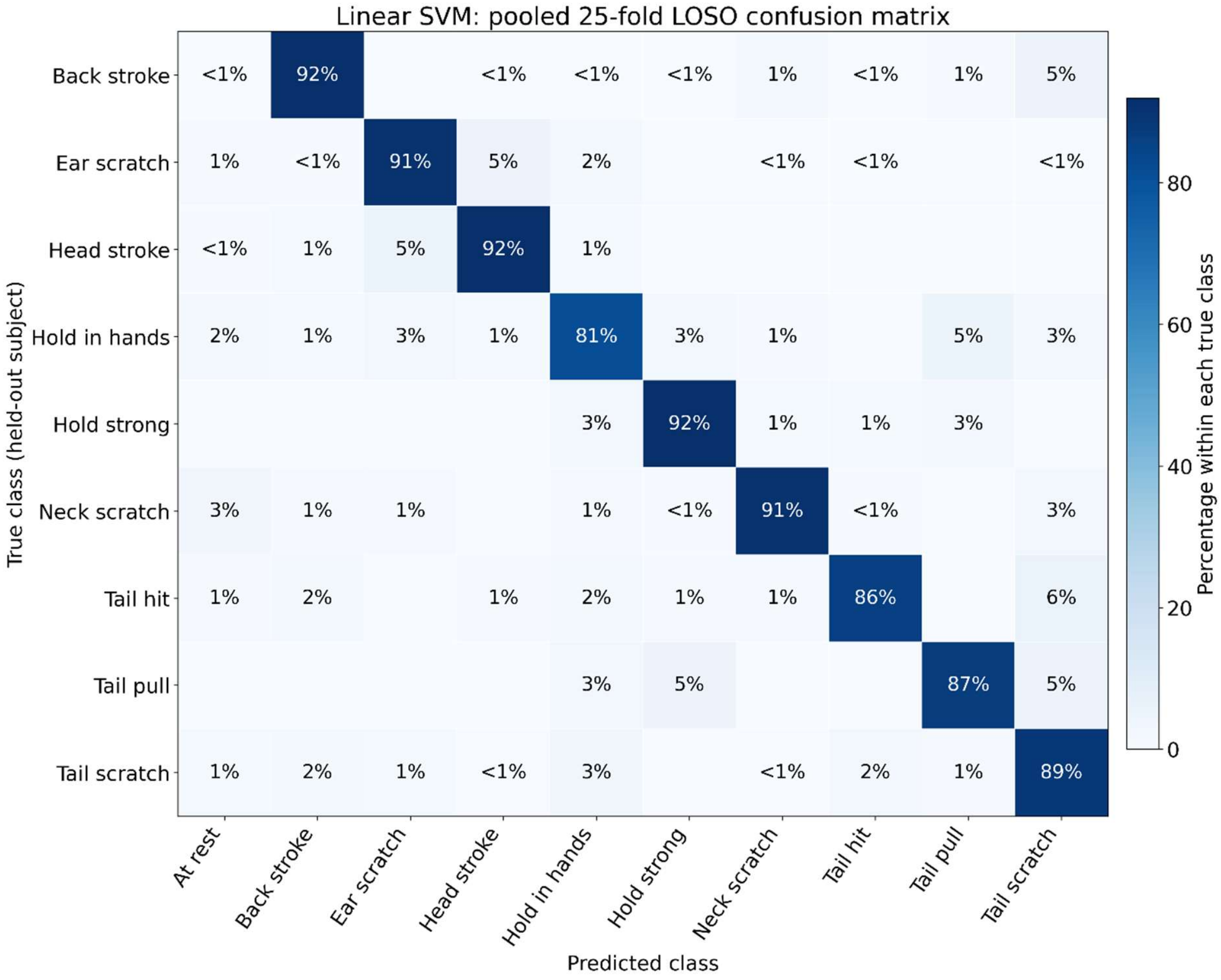

**Figure 11. Row-normalized pooled OOF confusion matrix for the linear SVM.**

The strong diagonal structure indicates that the linear SVM distinguished most operational gesture classes consistently across unseen participants. Class-specific recognition rates ranged from 81% to 92%. Emerging patterns indicate that the feature-based classifier captured the dominant differences between gestures, while its residual errors primarily involved interactions with similar sensor activation profiles or greater inter-participant variability.

### 4.6. ESP32-S3 Deployment Results

The compact dilated CNN, DS-CNN, TCN, linear SVM, and random forest achieved 100/100 host–board label agreement and were accepted as valid deployments. The compact Transformer achieved only 75/100 agreement, with the same disagreements in controlled single- and dual-core runs; its timing is therefore reported only as an invalid deployment observation and not used to recommend the model. The source of this disagreement remains unresolved. The identical prediction differences observed under single- and dual-core execution direct the working hypothesis toward a numerical compatibility issue introduced during quantized model conversion or within the ESP-DL implementation of one or more Transformer operators. Future operator-level parity analysis will localize the stage at which the host and embedded outputs begin to diverge. DS-CNN and TCN were substantially faster than the original dilated implementation because their strided structures reduced intermediate sequence lengths. The obtained measurements are presented in Table 8.

**Table 8. ESP32-S3 deployment size, parity, and measured latency.**

| Model | Precision | Serialized bytes | Model mean | End-to-end mean | Host–board agreement | Valid |
|---|---|---|---|---|---|---|
| Dilated CNN 14d1_39d2_41d4 (8–12–16) | W8A8 | 23,968 | 18.635 ms | 19.895 ms | 100/100 | Yes |
| DS-CNN | W8A8 | 28,896 | 3.940 ms | 5.177 ms | 100/100 | Yes |
| TCN | W8A8 | 47,056 | 4.263 ms | 5.969 ms | 100/100 | Yes |
| Compact Transformer | W8A8 | 39,312 | 7.341 ms | 8.820 ms | 75/100 | No |
| Linear SVM | FP32 | 3,296 | 0.219 ms | 1.405 ms | 100/100 | Yes |
| Random forest | FP32 | 238,096 | 1.453 ms | 2.556 ms | 100/100 | Yes |

For the SVM, the 1.405-ms end-to-end mean included approximately 1.080 ms for channel normalization, 0.350 ms for feature extraction, and 0.219 ms for prediction; the summed component means differ slightly from the directly timed total because the phases were instrumented separately. Neural model-context reports were approximately 203–209 KiB of internal RAM plus 15–34 KiB of PSRAM, depending on architecture. Static C++ coefficient arrays did not expose an equivalent runtime-context measure, so the reported zero context value for traditional models was not interpreted as zero working memory.

Raw-window preprocessing for the compact dilated CNN required a mean of 1.799 ms. Model execution required 18.419 ms, while the directly measured end-to-end path required 19.895 ms, with a p95 latency of 19.944 ms. Both isolated model execution and complete-path predictions agreed with the host implementation for all 100 validation windows.

The smallest shortlisted dilated CNN reached 14.637 ms mean model-only inference, but its mean macro-F1 was only 76.64%. The compact dilated CNN remained feasible for embedded execution, requiring 18.419 ms for model inference and 19.895 ms for the complete raw-window classification path. The latter represents approximately 4.0% of the 500 ms classification-update interval.

### 4.7. Formative Real-Time Observations

Selected classifiers were connected to the live sensor stream through a computer-based interface displaying sensor signals and class probabilities. A limited number of participants performed representative gestures while comparing the learned predictions with the responses of legacy heuristic algorithm.

The main qualitative improvement concerned subtle strokes and scratches, which produced low-amplitude changes distributed across multiple sensors and over time. The learned classifier frequently recognized these patterns below the heuristic activation level, resulting in visibly earlier and more sensitive responses. Abrupt high-force actions produced pronounced signal changes and activated the threshold logic promptly. This complementary behaviour motivated a hybrid design combining threshold detection for energetic events with learned classification for nuanced interactions. Because the activity lacked instrumented onset markers, blinded annotation, and a controlled participant protocol, no numerical accuracy or response-time claim is made. However, these observations provided sufficient formative evidence to motivate implementation and controlled real-world evaluation of the hybrid design as the next stage of development.

## 5. Discussion

### 5.1. From Raw-Signal CNN Design to Evidence-Based Model Selection

The study began with the hypothesis that a compact dilated 1D CNN could replace manual rules by learning useful temporal representations directly from minimally processed sensor windows. The MATLAB architecture study supported that hypothesis in a limited but important sense: it identified small networks that captured the complete interaction interval and produced useful subject-independent performance without handcrafted features. The compact CNN's 81.96% mean macro-F1, 23.4-kB quantized payload, exact host–board parity, and sub-20-ms model execution establish it as a viable embedded raw-window model.

The stricter comparison also showed that the initial wish to avoid feature engineering was not the most effective decision for this dataset. Six simple, physically interpretable summaries per channel allowed a linear SVM to achieve higher macro-F1, smaller model storage, and lower end-to-end latency. The convolutional investigation therefore establishes the complementary trade-off between automatic representation learning and feature-based efficiency: the CNN automates representation learning and reduces ongoing feature-maintenance effort, whereas the present small-sample sensor problem is sufficiently structured that carefully bounded time-domain features generalize better.

Only 25 statistically independent human contributors were available despite thousands of overlapping windows. Linear models can regularize effectively in that regime, and the selected features—level, variation, extrema, energy, and temporal roughness—map closely to the physical distinctions between holding, stroking, scratching, and impact. As future datasets grow and behaviour becomes more diverse, learned features may gain relative value; the present results should not be generalized beyond the studied sensor arrangement and taxonomy.

### 5.2. Interpretation of the 18-Class and 10-Class Tasks

The 18-class formulation was intentionally demanding. It retained table-versus-held contexts and closely related tail gestures to challenge receptive-field designs and expose ambiguities. That stress test was useful during model development, but preserving every distinction in the embedded decision output had no practical benefit once candidate architectures were identified. The 10-class mapping therefore reflects the response policy the toy must implement. The improvement across all CNNs indicates that several 18-class errors involved labels that were separable in annotation but not necessary for different behaviour.

### 5.3. Predictive Performance versus Embedded Cost

DS-CNN provided the most attractive validated neural alternative: it matched the compact dilated CNN in parameter count, used approximately 27% of its MACs, and achieved a slightly higher mean macro-F1. The paired difference was not significant after correction, and Figure 10 shows substantial overlap across participants, so the compact dilated CNN is not completely dominated as a predictive model. Nevertheless, the direct hardware result demonstrates that operation structure and memory access matter more than parameter count alone; the DS-CNN and TCN ran several times faster.

The compact Transformer illustrates why nominal parameter/MAC estimates are insufficient. Its predictive accuracy was valid, but 75% host–board label agreement failed the deployment gate. The model is therefore neither hidden nor recommended. The random forest achieved strong LOSO performance and short latency but required a 238-kB payload, whereas the linear SVM required only 3.2 kB. For the current system, the SVM is the most defensible primary classifier; DS-CNN is the strongest validated feature-free neural alternative, and the dilated CNN remains the central architecture studied and a useful baseline for raw-window inference.

### 5.4. MATLAB–Python–ESP32-S3 Workflow

The final workflow presented in the paper can be viewed as a progression. Processing and MATLAB support the initial labour-intensive work: guided recording, visual quality control, interactive trimming, window design, manual experiments, and large-scale CNN exploration. Python reproduces those steps with participant provenance and fold-local preprocessing, uses GPU acceleration for repeated LOSO training, and was created as a bridge to export models and artifacts suitable for ONNX, ESP-DL, and static C++ implementations. VS Code then provides the natural environment for the C++ benchmark and future firmware integration.

### 5.5. Implications for a Hybrid Interaction System

The results support a hybrid architecture in which threshold-based and learned decisions operate in parallel. Simple rules can respond immediately to high-energy events, while a classifier processes rolling multichannel windows to distinguish subtle, sustained, and spatially distributed gestures. An arbitration layer can combine event intensity with the predicted gesture class when selecting the toy's response.

The linear SVM is the leading candidate for initial integration because it achieved the highest mean LOSO accuracy and macro-F1, occupied 3.3 kB, and completed classification in 1.405 ms. The DS-CNN offers direct window-based neural processing with a 5.177 ms end-to-end latency.

The next stage will integrate the selected classifier with the heuristic path in the ESP32-S3 firmware.

## 6. Limitations

The dataset contains substantial manual collection effort and demographic diversity, but only 25 independent participants. The sample size was determined by practical feasibility rather than an a priori power analysis; subgroup performance by age or sex cannot be estimated reliably. The supervised session format, one physical toy design, one sensor arrangement, and one institutional context may not represent unconstrained home, classroom, or clinical interaction. Longer recordings, repeated sessions, device-to-device variation, sensor drift, material deformation, humidity, battery state, and adversarial or simultaneous contacts were not systematically studied.

The 2.5-s window was selected empirically rather than through a dedicated sensitivity experiment. Overlapping windows were kept within participant partitions, but they remain correlated observations; participant-level confidence intervals and paired tests reduce, but do not remove, the limitations of a small number of independent subjects. At-rest data were subjectless by design and were excluded from the primary participant-generalization metrics. The operational class merging was justified by response equivalence, yet different applications may require the original fine-grained distinctions.

The comparison was intentionally bounded to embedded-feasible models and modest tuning spaces. It does not include every TCN, TinyML architecture, recurrent network, Transformer, feature library, or ensemble. The compact Transformer result is not a general statement about attention models; it reflects one resource-matched architecture and a failed parity outcome in the tested ESP-DL toolchain. Similarly, the superior SVM result depends on six handcrafted features and may change with a larger or less structured dataset.

Hardware timing measured isolated classifier computation on the ESP32-S3 benchmark harness, not complete toy response latency including sensor buffering, decision smoothing, audio/LED control, and concurrent firmware tasks.

The PC live exercise was qualitative and cannot establish accuracy, robustness, or reaction-time superiority. Clinical, educational, and emotional outcomes are outside the scope of this technical study.

### 7. Conclusions and Future Work

This study developed and evaluated a complete path from supervised tactile data collection to measured embedded inference for a soft interactive companion. The original MATLAB workflow enabled acquisition, visual data curation, sliding-window preparation, and a 468-run investigation of compact dilated 1D CNNs. A deployment-oriented Python implementation preserved participant provenance, eliminated cross-participant leakage, fitted preprocessing within each fold, and connected the selected models to reproducible ESP32-S3 artifacts.

The 18-class task served as a deliberately high-granularity challenge; the response-aligned 10-class task provided the more suitable operational target. On identical 25-fold LOSO splits, the compact dilated CNN remained a viable feature-free raw-window classifier, while DS-CNN offered the strongest validated neural trade-off. The main empirical conclusion, however, is that a linear SVM using six interpretable features per channel achieved the best performance and deployment efficiency for the current data: 87.98% mean macro-F1, 91.05% mean fold accuracy, a 3,296-byte payload, 1.405-ms mean end-to-end classifier time, and exact host–board label agreement.

Future work will integrate the selected classifier with the heuristic path in the interactive toy, instrument the full response chain, and compare alternative decision policies in controlled yet realistic interactions.

### Data and Software Availability

The original dataset is openly available in Zenodo at https://doi.org/10.5281/zenodo.17650600 [54]. The original Processing/MATLAB/Python workflow is openly available in Zenodo at https://doi.org/10.5281/zenodo.17650545 [55].

### Glossary

| | |
|---|---|
| 1D CNN | — One-Dimensional Convolutional Neural Network |
| ASD | — Autism Spectrum Disorder |
| CNN | — Convolutional Neural Network |
| CSV | — Comma-Separated Values |
| DS-CNN | — Depthwise-Separable Convolutional Neural Network |
| FAIR | — Findable, Accessible, Interoperable, Reusable |
| FP32 | — 32-bit Floating Point |
| GUI | — Graphical User Interface |
| LOSO-CV | — Leave-One-Subject-Out Cross-Validation |
| MAC | — Multiply–Accumulate Operation |
| MCU | — Microcontroller Unit |
| ML | — Machine Learning |
| OOF | — Out-of-Fold |
| PSRAM | — Pseudo-Static Random-Access Memory |
| RF | — Random Forest |
| RMS | — Root Mean Square |
| SAR | — Socially Assistive Robot |
| SIMD | — Single Instruction Multiple Data |
| SVM | — Support Vector Machine |
| TCN | — Temporal Convolutional Network |
| W8A8 | — 8-bit Weights and 8-bit Activations |

## Author Contributions

**Aleksandrs Vališevskis:** Conceptualization, Methodology, Software, Writing-Original draft, Visualization, Investigation. **Aleksandrs Okss:** Conceptualization, Methodology, Writing-Review & Editing. **Inese Tīģere:** Methodology, Writing-Review & Editing. **Aleksejs Kataševs:** Conceptualization, Methodology. **Dina Bethere:** Methodology, Project administration. **Anete Hofmane:** Methodology, Validation. **Airisa Šteinberga:** Methodology, Validation. **Undīne Gavriļenko:** Methodology, Validation. **Santa Meļķe:** Validation. **Lucie Matoušková:** Validation.

## Acknowledgements

Funding: This work was supported by research and development grant No. RTU-PA-2024/1-0074 under the EU Recovery and Resilience Facility funded project No.5.2.1.1.i.0/2/24/I/CFLA/003 “Implementation of consolidation and management changes at Riga Technical University, Liepaja University, Rezekne Academy of Technology, Latvian Maritime Academy and Liepaja Maritime College for the progress towards excellence in higher education, science, and innovation” (RTUID 4835).